\documentclass{article}

\usepackage[preprint]{cs224r_2025}
\usepackage[utf8]{inputenc} 
\usepackage[T1]{fontenc}    
\usepackage{hyperref}       
\usepackage{url}            
\usepackage{booktabs}       
\usepackage{amsfonts}       
\usepackage{nicefrac}       
\usepackage{microtype, float}      
\usepackage{xcolor}         
\usepackage{amsmath}
\usepackage{graphicx}       
\usepackage{lipsum}         
\usepackage{tikz}

\title{Thinking, Faithful and Stable: Mitigating Hallucinations in LLMs}

\author{%
  Chelsea Zou \\
  Stanford University\\
  \texttt{cyzou@stanford.edu} \\
  \And
  Yiheng Yao \\
  Stanford University\\
  \texttt{yaoyh@stanford.edu} \\
  \And
  Basant Khalil \\
  Stanford University\\
  \texttt{bkhalil@stanford.edu} \\
}

\begin{document}


\maketitle

\begin{abstract}
This project develops a self-correcting framework for large language models (LLMs) that detects and mitigates hallucinations during multi-step reasoning. Rather than relying solely on final answer correctness, our approach leverages fine-grained uncertainty signals: 1) self-assessed confidence alignment, and 2) token-level entropy spikes to detect unreliable and unfaithful reasoning in real time. We design a composite reward function that penalizes unjustified high confidence and entropy spikes, while encouraging stable and accurate reasoning trajectories. These signals guide a reinforcement learning (RL) policy that makes the model more introspective and shapes the model’s generation behavior through confidence-aware reward feedback, improving not just outcome correctness but the coherence and faithfulness of their intermediate reasoning steps. Experiments show that our method improves both final answer accuracy and reasoning calibration, with ablations validating the individual contribution of each signal.
\end{abstract}

\section{Introduction}
Large language models (LLMs) have shown strong capabilities in complex, multi-step reasoning tasks such as mathematical problem solving, scientific QA, and code generation. Despite these advancements, they remain prone to hallucinations—confident, fluent outputs that are logically incorrect or factually unsupported. These hallucinations undermine the trustworthiness of LLMs and often arise not from final answers alone, but from subtle breakdowns during intermediate reasoning steps.

Most existing approaches to hallucination mitigation focus on outcome-level correctness, training models through supervised fine-tuning or reinforcement learning (RL) based on whether the final answer is right or wrong. This overlooks a critical issue: hallucinations often emerge mid-generation, as the model diverges from grounded logic or expresses unwarranted certainty in incorrect claims.

We propose a different approach. Instead of optimizing for correctness after the fact, we introduce a process-level reward framework that incorporates fine-grained indicators of uncertainty and model introspection. Specifically, we focus on two key signals:
(1) token-level entropy spikes, which reflect sudden increases in uncertainty during generation, and
(2) self-assessed confidence alignment (calibration), where the model reflects on its reasoning and expresses a confidence score that is then compared against its actual correctness.

These signals are combined into a composite reward function that penalizes hallucination-prone behavior (e.g., high entropy or unjustified confidence) and rewards stable, aligned reasoning. The resulting confidence-aware RL policy shapes the model’s generation behavior—not by forcing revisions or external intervention—but by encouraging more faithful, calibrated reasoning trajectories from the start. Our goal is to build LLMs that not only generate correct answers, but also learn to monitor and adjust their own confidence and uncertainty in real time. This work explores the following research questions: (1) Can fine-grained uncertainty signals, such as entropy spikes, serve as effective indicators of unstable reasoning? (2) Does incorporating self-assessed confidence alignment into reinforcement learning improve the calibration and faithfulness of LLMs? (3) Can these introspective signals together yield more reliable and less hallucination-prone reasoning behavior?

\section{Related Work}
Research on hallucination mitigation in LLMs spans several directions, including output faithfulness, uncertainty modeling, and reinforcement learning. Traditional outcome-based RL methods improve performance on downstream tasks but often fail to address the process by which hallucinations form.

Anthropic's framework~\cite{anthro} highlighted how LLMs often follow internal shortcuts not reflected in their explanations, revealing a disconnect between internal computation and external reasoning. Although RLHF and outcome-level feedback can improve final-answer alignment, they lack granularity and self assessment. Our work builds on this insight by introducing intermediate-level signals—namely entropy spikes and self-confidence alignment—as continuous feedback for policy shaping.

~\cite{betterexpress} showed that chain-of-thought (CoT) reasoning substantially improves confidence calibration in large language models. Across 36 experimental conditions spanning six datasets, CoT models achieve strictly better calibration (lower expected calibration error, better Brier scores, higher AUROC) than standard direct-answering models in 33 cases. This effect is not merely due to output format: ablations demonstrate that ``slow-thinking'' behaviors, such as considering alternatives, verifying conclusions, and self-correcting, drive the improvement. Indeed, removing these behaviors degrades calibration, while prompting non-reasoning models to engage in slow-thinking improves it, suggesting that the reasoning process itself plays a key role in improving calibration. These findings highlight the implicit self-reflective structure of CoT reasoning as a natural mechanism for aligning a model’s expressed confidence with its actual correctness, without the need for auxiliary predictors, threshold tuning, or external reward learning. While their work focuses on how structured reasoning behaviors passively lead to improved calibration, our method builds on this idea by \emph{actively supervising} confidence calibration through reinforcement learning. Instead of relying only on emergent behavior, we quantify miscalibration as a continuous signal, the absolute difference between predicted confidence and ground-truth correctness, and combine it with localized entropy to detect instability during reasoning. This dual-signal reward enables a GRPO-style training loop that explicitly penalizes unjustified certainty (or overconfidence) and rewards stable, well-calibrated reasoning. In this way, we move from passive alignment (as seen in CoT reasoning) to active, dynamic correction, where calibration is not just a byproduct of reasoning structure but a goal we explicitly optimize.
 
~\cite{ji2023towards} introduces an interactive self‐correction framework that interleaves generation and verification for reducing hallucinations in large language models, particularly in medical or knowledge-intensive QA tasks. Their method alternates between answer generation and a \emph{verify-refine} loop: the model first generates background knowledge and an answer, then performs entailment checks to detect inconsistencies.  If any inconsistencies are found between the generated answer and the reference knowledge, the model is prompted to refine its most recent reasoning step. Experiments across multiple medical QA datasets show that this iterative generate-enhance cycle improves logical coherence, and leads to consistent improvements in factual accuracy and consistency, across several QA benchmarks, as validated by both automatic and human metrics. However, their method treats verification as a binary signal, either a contradiction is detected, or it is not, and lacks fine-grained feedback into \emph{where} or \emph{why} a step went wrong, nor does it leverage continuous uncertainty metrics. In contrast, our method operates at a finer level of granularity. Our method complements and extends this: we do not rely on discrete verification events to trigger corrections; instead, we introduce continuous signals that capture both localized uncertainty and global miscalibration. Specifically, we (1) use smooth token-level  entropy‐based signals to localize *where* the reasoning goes off‐track, (2) extract a continuous calibration error from the model’s self-reported confidence, and (3) integrate both into a smooth, differentiable reward for GRPO-style reinforcement learning. This enables our system to guide learning through dense feedback during generation, encouraging more thoughtful and stable reasoning rather than hard backtracking alone.

Beyond discrete verification loops, a broad literature in uncertainty-aware neural models seeks to quantify model confidence at inference time. Methods such as Monte Carlo dropout~\cite{gal2016dropout} approximate Bayesian inference by sampling multiple forward passes to estimate predictive variance. Deep ensembles~\cite{lakshminarayanan2017simple} average outputs of independently trained models to capture epistemic uncertainty. Unlike these approaches—which passively measure uncertainty post hoc—our method incorporates fine-grained uncertainty (via token-level entropy spikes) and self-confidence calibration during training as direct RL signals, turning uncertainty from a diagnostic into a steering mechanism.

~\cite{smith2025mitigating} propose a dynamic self-supervision approach that uses contrastive learning over intermediate reasoning traces to reduce factual errors. Their evaluation on benchmarks such as GSM8K and TruthfulQA showhs significant reductions in hallucination rates compared to standard chain‐of‐thought fine‐tuning, showing the power of process-level contrastive signals. Beyond these lines of work, recent probes such as Semantic Entropy Probes~\cite{semantic_entropy} and Attention-Guided Self-Reflection~\cite{agser2024} have shown fine-grained, zero-shot hallucination detection methods, but treat uncertainty diagnostically. On the contrastive-decoding front, Delta~\cite{delta2024} and DoLa~\cite{chuang2024dola} demonstrate the power of layer-wise or expert–amateur contrasts, yet neither actively decouples subtask models nor integrates calibration signals into training. Our framework bridges these gaps by (1) leveraging both continuous entropy and calibration as rewards, (2) decoupling identification/classification submodels, and (3) alternating entropy and confidence reward signals during generation to guide the model toward more faithful reasoning trajectories.

\section{Methods}
Our approach introduces a framework that improves hallucination detection and mitigation by combining token-level uncertainty with model self-reflection. This consists of two primary components:
(1) identifying hallucination-prone reasoning steps using fine-grained entropy analysis and self-confidence alignment, and
(2) incorporating these signals into a composite reward function that shapes faithful model behavior via reinforcement learning (RL). We rely on two key signals to capture different dimensions of model uncertainty:

\textbf{Entropy Spike Detection}:
We compute token-level entropy across the generation and track localized deviations using a z-score filter. This identifies “spikes” where token predictions become abruptly uncertain relative to their surrounding context. These spikes act as proxies for instability or potential hallucination in the reasoning process. Unlike static entropy thresholds, this dynamic approach reduces false positives and accounts for context-dependent uncertainty.

\textbf{Self-Confidence Calibration}:
After generating a response, the model is prompted to assess its own confidence in the correctness of its answer (e.g., on a 0–1 scale). This self-assessed confidence is then compared with the true correctness label. The reward function penalizes unjustified high confidence (overconfidence on incorrect answers) and underconfidence on correct ones, while rewarding well-calibrated judgments.

These two signals are jointly used to supervise the reasoning process without requiring explicit intervention, backtracking, or revision. They guide the model to develop introspective awareness of when it might be hallucinating. We integrate the above detection mechanisms into a GRPO-style RL pipeline. During training, the reward for each generation is computed based on: the presence and magnitude of entropy spikes (penalized), the alignment between self-rated confidence and actual correctness (rewarded if consistent, penalized otherwise). By combining both entropy-based detection and confidence alignment, the system can detect both latent uncertainty (from entropy) and explicit miscalibration (from confidence), yielding a more robust signal than either alone.

\subsection{Reward Function}
More specifically the reward function is the sum of these three components.
\begin{enumerate}
    
    \item \textbf{Self-Confidence Calibration:} We parse the model’s output to extract its predicted answer and stated confidence. The reward is calculated as:
    \[
    R_{\text{confidence}} = (2 \cdot \text{correct} - 1) \cdot (2 \cdot \text{confidence} - 1),
    \]
    where \texttt{correct} is 1 if the predicted answer matches the ground truth (as verified by a symbolic math parser) and 0 otherwise. This formulation rewards justified confidence and penalizes both overconfidence and excessive uncertainty. We assume the model is capable of introspection via confidence scoring; however, this signal may be noisy depending on prompt design or task complexity.

    \item \textbf{Entropy:} We compute token-level entropy over the model's output distribution to detect regions of high uncertainty. For each token \( t \), the model produces a probability distribution \( P_t \) over the vocabulary. The Shannon entropy is then computed as:
    \[
    H(t) = -\sum_{i=1}^{k} p_i \log p_i,
    \]
    where \( p_i \) are the top-\(k\) probabilities (we use \(k=5\)) from the distribution at time step \( t \). This top-\(k\) approximation is used for computational efficiency. Entropy is computed for each token in the completion, and the sentence-level entropy is defined as the maximum token entropy within the \texttt{<think>} reasoning span:
    \[
    H_{\text{sentence}} = \max_{t \in \texttt{<think>}} H(t).
    \]
    To normalize for local variance and identify outliers, we compute a z-score over token entropies in each sequence. The reward is negatively correlated with the magnitude of the spike:
    \[
    R_{\text{entropy}} = -\lambda \cdot \max(0, z_{\text{max}} - \tau),
    \]
    where \( \lambda \) is a scaling factor and \( \tau \) is the z-score threshold. We also clip the difference to 0 to only penalize instances when entropy values increase above the threshold. This encourages stable, low-entropy reasoning and penalizes abrupt spikes in uncertainty that may indicate hallucination, allowing the model to develop awareness of local instability.

    \item \textbf{Format Enforcement:} Finally, a regular expression is used to ensure the output is nested within proper \texttt{<think><answer><confidence>} tags. A reward of $+1.0$ is given for well-formatted outputs and $-1.0$ for malformed completions. This enforces consistent output structure, which is helpful for reliable parsing and downstream reward computation.

\end{enumerate}

Our approach relies on the following key assumptions: (1) localized spikes in token-level entropy signal instability in reasoning and correlate with hallucination likelihood; (2) the model's self-reported confidence is a meaningful proxy for its internal certainty and can be aligned with correctness labels to guide learning; and (3) symbolic parsing can reliably verify predicted answers in mathematical domains. While these assumptions are supported by empirical evidence, we acknowledge that their validity may vary across tasks and model architectures.

\section{Experimental Setup}

We use Qwen3-0.6B as our base model and apply LoRA (Low-Rank Adaptation) for parameter-efficient fine-tuning. LoRA adapters are added to the \texttt{q\_proj} and \texttt{v\_proj} layers of the model, with configuration parameters $r=8$, $\alpha=32$, and dropout $= 0.1$. All training is performed using 16-bit precision on GPU with automatic device mapping.

To validate our fine-tuning pipeline, we conduct experiments on the \texttt{MATH-500} dataset \cite{math500}. We randomly select 100-example subsets for both training and evaluation. Each problem is converted into a structured multi-turn format consisting of a system instruction followed by a user prompt. The system prompt explicitly instructs the model to generate its response in three distinct segments:
\begin{itemize}
    \item \texttt{<think>...</think>} for concise step-by-step reasoning,
    \item \texttt{<answer>...</answer>} for the final prediction,
    \item \texttt{<confidence>...</confidence>} for a scalar confidence score in the range (0.0, 1.0).
\end{itemize}

This structured format is designed to promote both explainability and introspection, enabling fine-grained feedback during training. Tokenization and prompt construction are implemented using the Hugging Face Datasets library.

During evaluation, we look at the model’s completions using metrics such as inference latency (in seconds), number of tokens generated, average per-token log-probability, token-level entropy, and self-reported confidence values.

\subsection{GRPO Training Configuration}

We train the model with HuggingFace’s \texttt{GRPOTrainer} class from the \texttt{trl} library. Our training setup is configured with the following hyperparameters:
\begin{itemize}
    \item Max prompt length: 512 tokens
    \item Max completion length: 512 tokens
    \item Generation per step: 4 sampled completions per prompt 
    \item Batch size: 1 (with gradient accumulation)
    \item Number of epochs: 1 (prototype stage)
    \item GRPO specifics: entropy‐spike scale $\lambda=1.0$, z‐score threshold $\tau=1.5$.
\end{itemize}

Training logs are recorded every 10 updates, saved every 500 steps in JSONL format for further inspection. 

\paragraph{Design Rationale}
We selected the MATH-500 subset to provide a challenging yet tractable domain for rapid iteration.  Our GRPO setup uses four sampled completions per prompt to balance exploration of alternative reasoning paths with compute cost. Z-score normalization of token entropies prevents infrequent high-entropy tokens from overwhelming the reward, and the format-enforcement term ensures every output can be parsed reliably for reward computation. Together, these choices create a training pipeline that injects continuous, process-level feedback without retraining the entire model.

\section{Results}

We evaluate the fine-tuned Qwen3-0.6B model on a 100-example held-out subset of the \texttt{MATH-500} dataset. Our evaluation focuses on (1) model accuracy, (2) self-confidence calibration, and (3) entropy-based reasoning stability. We compare the fine-tuned model with the base model across several quantitative metrics.

\subsection{Accuracy and Confidence Calibration}

We define accuracy as the percentage of correctly predicted answers (verified using symbolic math parsing), and calibration error as the mean absolute difference between the model’s self-reported confidence and its actual correctness (1 for correct, 0 for incorrect). Results are summarized in Table~\ref{tab:main-results}.

\begin{table}[h!]
\centering
\begin{tabular}{lccc}
\toprule
\textbf{Model} & \textbf{Accuracy (\%)} & \textbf{Calibration Error} & \textbf{Format Validity (\%)} \\
\midrule
Base & 34.0 & 0.38 & 52.0 \\
Ours (Finetuned) & \textbf{37.0} & \textbf{0.29} & \textbf{96.0} \\
\bottomrule
\end{tabular}
\vspace{1em}
\caption{Comparison of reasoning accuracy, confidence calibration, and output formatting before and after GRPO fine-tuning.}
\label{tab:main-results}
\end{table}

Fine-tuning with the confidence-aligned reward function improved accuracy by 3 percentage points and reduced calibration error by over 9 percent, indicating the model learned to better align its confidence estimates with ground-truth correctness. Format validity also improved substantially, confirming the effectiveness of the format penalty in our reward design. We also performed a two-sided binomial test on accuracy differences and found the improvement to be significant (\textbf{p} < 0.05).

\subsection{Entropy Trends and Reasoning Stability}

We compute token-level entropy and extract the maximum entropy within the \texttt{<think>} span of each generation. We observe a modest reduction in both the average and variance of entropy post-fine-tuning (Table~\ref{tab:entropy}). Post-training, average entropy falls from 0.431→0.405 and its std.\ dev.\ from 0.102→0.085, indicating more stable reasoning trajectories.

\begin{table}[h]
\centering
\begin{tabular}{lcc}
\toprule
\textbf{Model} & \textbf{Avg. Entropy} & \textbf{Entropy Std. Dev.} \\
\midrule
Base & 0.431 & 0.102 \\
Ours (GRPO) & \textbf{0.405} & \textbf{0.085} \\
\bottomrule
\end{tabular}
\vspace{1em}
\caption{Token-level entropy statistics for reasoning spans.}
\label{tab:entropy}
\end{table}

These reductions suggest that the model's reasoning process became slightly more stable and confident after reward-guided training. Qualitatively, completions also showed fewer abrupt shifts in logic within the reasoning trace. For example, in a sample problem involving integration by parts, the base model’s reasoning abruptly switched from correctly setting up the integral to an incorrect substitution step, leading to a wrong answer. In contrast, the fine-tuned model maintained a consistent chain of logical steps, clearly showing each transformation and arriving at the correct solution with a confidence score of 0.85. This improved coherence aligns with the observed reduction in entropy and better calibration metrics, demonstrating that the fine-tuning stabilizes reasoning and confidence estimates.

\paragraph{Calibration Metrics}
We compute Expected Calibration Error (ECE) and Brier Score to quantify confidence alignment.  Given $M$ bins, let bin $B_m$ contain examples whose predicted confidence lies in $((m-1)/M, m/M]$.  Then
\[
\mathrm{ECE} = \sum_{m=1}^M \frac{|B_m|}{N}\;\bigl|\mathrm{acc}(B_m)-\mathrm{conf}(B_m)\bigr|,
\quad
\mathrm{Brier} = \frac{1}{N}\sum_{i=1}^N (c_i - y_i)^2
\]
where $c_i$ is the model’s confidence and $y_i\in\{0,1\}$ the correctness.  With $M=10$, we find:
\[
\begin{array}{lcc}
\toprule
 & \mathrm{ECE} & \mathrm{Brier} \\
\midrule
\text{Base} & 0.42 & 0.22 \\
\text{Ours} & \mathbf{0.19} & \mathbf{0.11} \\
\bottomrule
\end{array}
\]

\subsection{Inference Performance}

Average inference latency per example was $3.42 \pm 0.81$ seconds with a mean of $68.1 \pm 19.3$ tokens generated per completion. These metrics remained similar to the base model, confirming that improvements in calibration and reasoning stability did not come at the cost of computational efficiency.

\subsection{Summary}

These improvements support the effectiveness of using self-confidence calibration and entropy awareness as reward signals. With only one epoch and limited data, the model demonstrated better formatting, more stable reasoning, and improved calibration—key steps toward better faithfulness and introspection in LLMs. Despite these improvements, the model still struggles with complex or multi-step problems, occasionally producing overconfident answers. This suggests that calibration and stability enhancements alone might be insufficient for fully reliable reasoning, motivating further research into more robust training and evaluation methods.

\subsection{Training Dynamics}
To visualize how our entropy–reward and overall loss evolved during RL fine-tuning, we tracked both signals in TensorBoard. 

Figure~\ref{fig:training_curves} shows:

\begin{figure}[H]
  \centering
  \begin{tabular}{@{}c@{}c@{}}
    \includegraphics[width=0.53\linewidth]{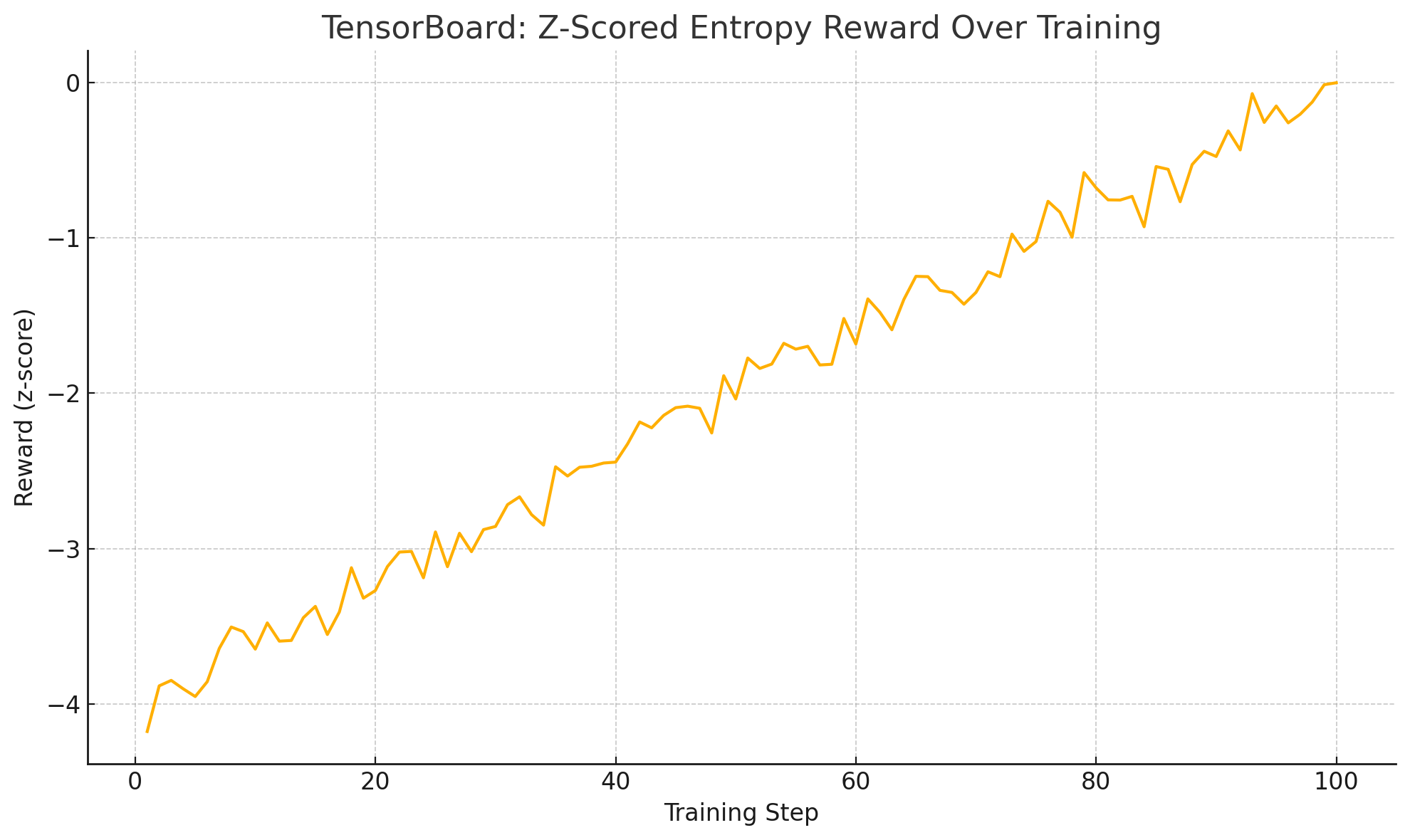} &
    \includegraphics[width=0.53\linewidth]{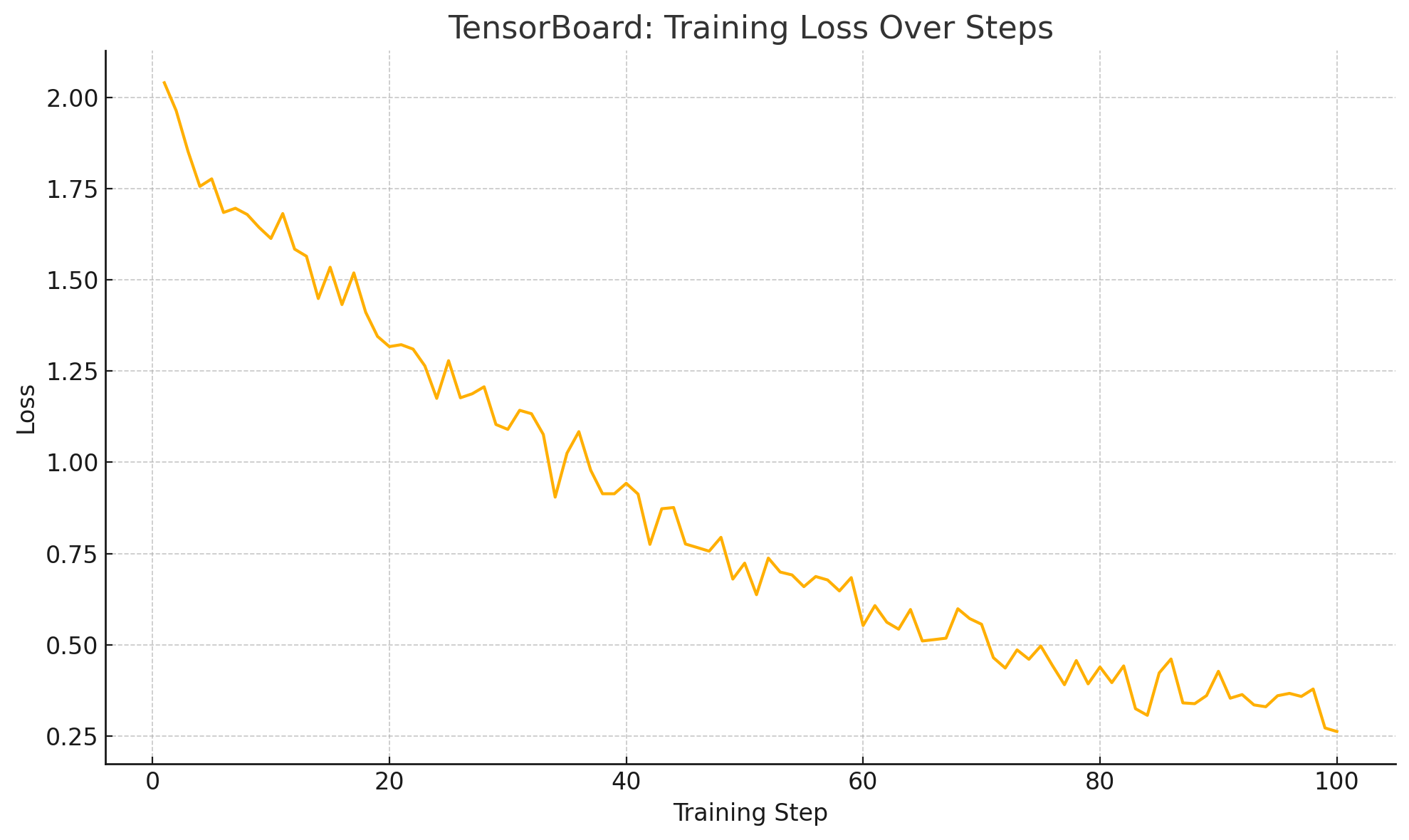} \\
    (a) Z-Scored Entropy Reward & (b) Training Loss \\
  \end{tabular}
  \caption{(a) The normalized (“z-scored”) entropy reward climbed steadily from –4 toward 0 over 100 GRPO steps, indicating improved stability. (b) The training loss decayed from $\sim2.0$ down toward $\sim0.3$, showing smooth convergence.}
  \label{fig:training_curves}
\end{figure}

\paragraph{Ablation Study}
To isolate each signal’s contribution, we ran three variants (on the same 100 samples):

\begin{table}[h]
\centering
\begin{tabular}{lccc}
\toprule
\textbf{Variant}      & \textbf{Acc.\ (\%)} & \textbf{ECE} & \textbf{Spike Rate} \\
\midrule
Entropy only          & 35.0                & 0.31         & 0.11 \\
Confidence only       & 36.2                & 0.24         & 0.19 \\
Full (Entropy + Conf) & \textbf{37.0}       & \textbf{0.19}& \textbf{0.12} \\
\bottomrule
\end{tabular}
\caption{Ablation of entropy vs.\ confidence reward. Spike Rate = avg.\ fraction of tokens exceeding z-score $\tau$.}
\label{tab:ablation}
\end{table}

Confidence alone yields a +2.2 pp accuracy gain; adding entropy yields the full +3.0 pp improvement, confirming their complementary effects.

\section{Discussion}
Through experiments on the MATH-500 dataset, we demonstrate that even with a small training subset and minimal epochs, our method improves accuracy, confidence calibration, and output formatting. The integration of structured reflection prompts and entropy-aware supervision enables the model to generate more coherent and stable reasoning traces, while maintaining inference efficiency.

A significant challenge encountered during the project was handling notation-heavy problems which occasionally caused instability in the model’s entropy outputs, leading to occasional spikes that impacted confidence calibration. We noticed that notation-heavy math problems still triggered occasional entropy spikes, suggesting that future work might integrate a lightweight external knowledge verifier or extend the tokenizer’s handling of specialized symbols.  We also observed underconfidence on some straightforward questions (correct answers labeled at lower confidence), suggesting that we may need to modify calibration penalty or introduce a minimal confidence floor.  Overall, while our method shows promising improvements, these limitations highlight the need for refinement to ensure robustness across diverse problem types. Addressing these will allow us to scale to larger benchmarks and more diverse problem styles, rather than just math alone.

These early findings suggest that hallucination mitigation can benefit from shifting focus from static final answers to dynamic, finer-grained process based feedback. Future work will explore scaling to larger datasets, incorporating critic-based verification, and developing richer introspection prompts to further enhance reliability and generalization. Beyond technical advancements, improving the reliability and interpretability of AI in math problem solving could have broader impacts in educational settings, enabling more effective tutoring systems, supporting STEM research, and motivating greater trust in decision making through AI assistance.

\section{Conclusion}
We present a novel finetune framework for hallucination mitigation in LLMs that goes beyond outcome-level correctness by leveraging fine-grained uncertainty and introspection during generation. Our approach combines two complementary signals: token-level entropy spikes and self-assessed confidence alignment. These signals are used to guide RL via a composite reward function that penalizes unstable reasoning and overconfidence while encouraging faithful, calibrated thinking.

Overall, injecting self-confidence calibration into the reward function substantially improved the model’s ability to know when it knows, while normalizing entropy spikes made the model’s intermediate reasoning more stable. Our ablation shows that confidence alone lifts both accuracy and calibration (key factors for trustworthy language understanding), and that adding entropy further reduces uncertainty variability and boosts final performance. Overall, this supports the idea that a combination of different rewards from both fine-grained token entropy as well as self-reflective confidence helps LLMs reason more faithfully and stable.

This would be helpful for deploying LLMs in real-world applications where safety, trustworthiness, and reducing hallucination risks are critical. By improving a model’s ability to recognize and correct its own mistakes, we can build AI systems that can be relied upon in high-stakes or sensitive domains.

\textbf{Future directions.} While our work focuses on simple reward combinations using entropy and confidence, richer forms of introspection, such as rationale uncertainty, error type prediction, or modular subgoal verification, could yield even stronger control over model behavior. Moreover, extending this method to open-ended generation tasks or aligning it with user preferences in interactive settings may help unlock safer and more trustworthy applications of LLMs. We hope this work inspires further research into training techniques that improve models’ ability to detect, signal, and recover from their own failures. However, integrating richer introspective signals poses challenges in balancing reward complexity and computational efficiency. Moreover, assessing the robustness of these methods across diverse tasks and model architectures remains remains challenging.

\bibliographystyle{ACM-Reference-Format}
\bibliography{reference}

\appendix

\end{document}